\documentclass[10pt,twocolumn,letterpaper]{article}

\usepackage{iccv}              

\usepackage{multirow}

\definecolor{iccvblue}{rgb}{0.21,0.49,0.74}
\usepackage[pagebackref,breaklinks,colorlinks,allcolors=iccvblue]{hyperref}

\def\confName{ICCV}
\def\confYear{2025}

\title{MVTrajecter: Multi-View Pedestrian Tracking \\ 
with Trajectory Motion Cost and Trajectory Appearance Cost}

\author{Taiga Yamane \and Ryo Masumura \and Satoshi Suzuki \and Shota Orihashi \and NTT Human Informatics Laboratries, NTT Corporation\\
{\tt\small \{taiga.yamane, ryo.masumura, satoshixv.suzuki, shota.orihashi\}@ntt.com}
}

\begin{document}
\maketitle
\begin{abstract}

Multi-View Pedestrian Tracking (MVPT) aims to track pedestrians in the form of a bird's eye view occupancy map from multi-view videos.
End-to-end methods that detect and associate pedestrians within one model have shown great progress in MVPT.
The motion and appearance information of pedestrians is important for the association, but previous end-to-end MVPT methods rely only on the current and its single adjacent past timestamp, discarding the past trajectories before that.
This paper proposes a novel end-to-end MVPT method called Multi-View Trajectory Tracker (MVTrajecter) that utilizes information from multiple timestamps in past trajectories for robust association.
MVTrajecter introduces trajectory motion cost and trajectory appearance cost to effectively incorporate motion and appearance information, respectively.
These costs calculate which pedestrians at the current and each past timestamp are likely identical based on the information between those timestamps.
Even if a current pedestrian could be associated with a false pedestrian at some past timestamp, these costs enable the model to associate that current pedestrian with the correct past trajectory based on other past timestamps.
In addition, MVTrajecter effectively captures the relationships between multiple timestamps leveraging the attention mechanism.
Extensive experiments demonstrate the effectiveness of each component in MVTrajecter and show that it outperforms the previous state-of-the-art methods.

\end{abstract}    

\begin{figure}[tb]
    \centering
    \includegraphics[width=80mm]{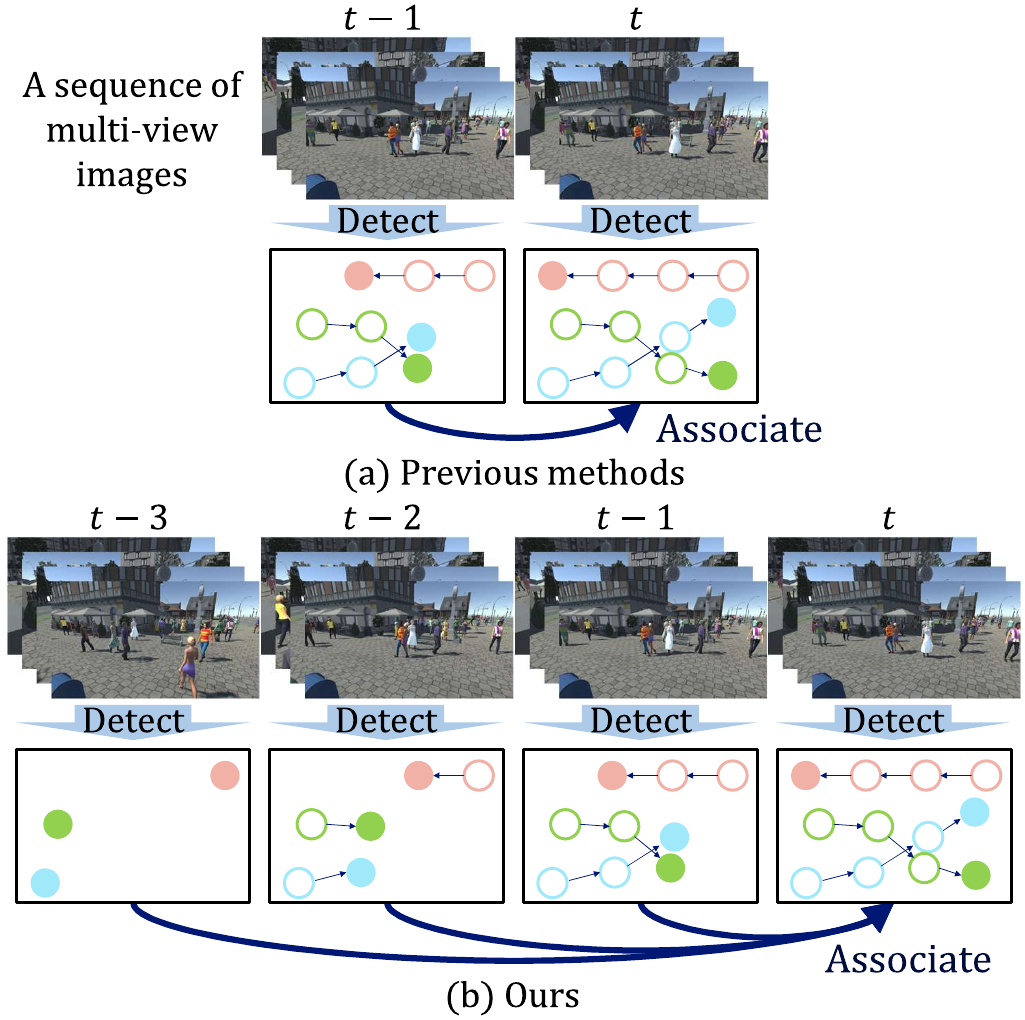}
    \vspace{-2mm}
    \caption{Comparison of (a) previous end-to-end MVPT methods and (b) MVTrajecter. This example shows a case where three pedestrians are on BEV occupancy maps and we want to associate pedestrians at $t$ with past pedestrians. The same color indicates the same pedestrian. The filled circles indicate detected pedestrians at each timestamp, and the empty circles indicate past pedestrians.}
    \vspace{-3mm}
    \label{fig:fig1}
\end{figure}

\section{Introduction}
\label{sec:intro}

Pedestrian tracking in videos is a crucial component in various applications, such as surveillance systems~\cite{elhoseny2020multi}, sports analysis~\cite{cui2023sportsmot}, and human-computer interaction~\cite{wengefeld2019multi}.
In pedestrian tracking, occlusion is a significant problem because it hampers tracking pedestrians hidden behind obstacles or other pedestrians.
As a potential solution, many studies have focused on \textit{Multi-View Pedestrian Tracking}~(MVPT)~\cite{chavdarova2018wildtrack,hou2020multiview,engilberge2023multi,ferryman2009pets2009,xu2016multi,fleuret2007multicamera}.
MVPT aims to track pedestrians in the form of a bird's eye view~(BEV) pedestrian occupancy map from multi-view videos captured by multiple cameras.
In contrast to monocular tracking~\cite{milan2016mot16,dendorfer2020mot20,sun2022dancetrack}, which uses a single non-calibrated camera, MVPT utilizes multiple calibrated cameras with partially overlapping fields of view.
MVPT is, therefore, expected to be more robust to occlusion in the overlapping area than monocular tracking by combining complementary clues.

A variety of MVPT methods have been proposed.
Early methods~\cite{berclaz2006robust,chavdarova2018wildtrack,ong2020bayesian,you2020realtime,nguyen2022lmgp,cheng2023rest,hofmann2013hypergraphs,eshel2008homography, hu2006principal,xu2016multi,specker2024ocmctrack,cherdchusakulchai2024online,quach2021dyglip,niculescu2024mctr,wang2024bev,qian2024bird,feng2024unveiling,han2024benchmarking,han2021multiple} relied on monocular object detection~\cite{ren2015faster,zhou2019objects}, which uses the detection results of each view image by monocular detection to localize pedestrians in the BEV occupancy map and identify which pedestrians are identical at different timestamps.
While these methods were important for early attempts in MVPT, occlusion in monocular detection hurt the final tracking performance.
Furthermore, these methods optimized monocular detection and subsequent parts separately, resulting in suboptimal tracking performances. 
To overcome these problems, end-to-end trainable methods~\cite{teepe2024earlybird,teepe2024lifting,engilberge2023multi,yang2024end,dakic2024resource} that do not rely on monocular detection have become the dominant approach.
These methods first generate a representation of the ground plane, called the BEV feature, for each timestamp and then track pedestrians through the detection and association processes using this feature.
In the detection process, they directly predict a BEV pedestrian occupancy map from the BEV feature for each timestamp.
In the association process, they identify which pedestrians are identical at the current and past timestamps by utilizing information on pedestrians extracted from the BEV feature.
These methods are optimized in an end-to-end manner.

In end-to-end MVPT methods, improving the association process has been the main research topic to achieve better tracking performances.
Previous methods utilize motion and/or appearance information of pedestrians between the current and an adjacent past timestamp.
EarlyBird~\cite{teepe2024earlybird} and REMP~\cite{dakic2024resource} extract appearance information from the BEV feature at each timestamp and extract motion information using a Kalman filter~\cite{kalman1960new}.
TrackTacular~\cite{teepe2024lifting}, MVFlow~\cite{engilberge2023multi}, and MVTr~\cite{yang2024end} concatenate the BEV features at the current and an adjacent past timestamp and extract motion information between those two timestamps.
Then, they associate pedestrians at the current and an adjacent past timestamp based on the appearance similarity and/or the distance between locations estimated by motions and actual detected locations.
While these methods have outperformed non-end-to-end MVPT methods, they only utilize the information on pedestrians between the current and an adjacent past timestamp and discard the information on past trajectories\footnote{Trajectories are parts of the final track and typically consist of more than two timestamps.} before that, as shown in Fig.~\ref{fig:fig1}(a).
Therefore, they do not fully utilize past trajectories, leading to association errors when the information at an adjacent past timestamp is unreliable (\eg, when an area is excessively crowded with pedestrians).
Although many end-to-end monocular tracking methods~\cite{zhou2022global,qin2023motiontrack,wang2020towards,yu2022towards,du2021giaotracker,du2023strongsort} have investigated leveraging information at multiple timestamps in past trajectories (see Sec.~\ref{sec:related}), it is not clear how to leverage this information in the end-to-end MVPT method.

In this paper, we propose a novel end-to-end MVPT method called \textbf{M}ulti-\textbf{V}iew \textbf{Traject}ory Track\textbf{er} (MVTrajecter), which performs the association based on motion and appearance information between the current and multiple timestamps in past trajectories, as shown in Fig.~\ref{fig:fig1}(b).
To effectively utilize the information, we introduce trajectory motion cost~(TMC) and trajectory appearance cost~(TAC).
These costs comprehensively identify which current pedestrian and which past trajectory are identical by calculating which current pedestrian is likely identical to which past pedestrian for each past timestamp.
Specifically, TMC calculates the distance between current locations estimated by motion information from each past timestamp and actual detected current locations, and TAC calculates the appearance similarities between pedestrians at the current and each past timestamp.
Even if a current pedestrian could be associated with a false pedestrian at some past timestamp, these costs enable MVTrajecter to associate that current pedestrian with the correct past trajectory based on the information at other past timestamps.
Furthermore, in order for TMC and TAC to identify pedestrians accurately, it is necessary to appropriately capture the relationships of pedestrians between multiple timestamps and extract identifiable motion and appearance information.
As shown in Fig.~\ref{fig:fig2}, MVTrajecter leverages the attention mechanism~\cite{vaswani2017attention}, which has been shown to have such ability.
During training, MVTrajecter learns to capture relationships of pedestrians between multiple timestamps through end-to-end training with TMC and TAC as its training objectives.
In experiments on three major MVPT datasets~\cite{chavdarova2018wildtrack,hou2020multiview,vora2023bringing}, we verify the effectiveness of MVTrajecter and demonstrate that utilizing information at multiple timestamps in past trajectories leads to a better tracking performance.

Our contributions are summarized as follows:
\begin{enumerate}[1.]
    \item We propose a novel end-to-end MVPT method called MVTrajecter. This comprehensively identifies which current pedestrian and which past trajectory is identical utilizing information between the current and multiple timestamps in past trajectories through TMC and TAC.
    \item Given the use of TMC and TAC, we design the architecture of MVTrajecter leveraging the attention mechanism and effectively capture the relationships between the current pedestrians and past trajectories, extracting identifiable motion and appearance information.
    \item We demonstrate that utilizing information at multiple timestamps in past trajectories leads to a better tracking performance. MVTrajecter outperforms previous methods and achieves a new state-of-the-art performance on GMVD, Wildtrack, and MultiviewX.
\end{enumerate}

\section{Related Work}
\label{sec:related}

\noindent
\textbf{Multi-View Pedestrian Detection.}
Multi-view pedestrian detection (MVPD)~\cite{chavdarova2018wildtrack,hou2020multiview,zhang2021cross,vora2023bringing} is a task to predict a BEV pedestrian occupancy map from a multi-view image.
Unlike MVPT, MVPD aims to detect pedestrians rather than track pedestrians and processes the multi-view image rather than the multi-view video.
MVPD is an important component of MVPT methods because they solve MVPT by detecting pedestrians at each timestamp and associating them across different timestamps (as described in Sec.~\ref{sec:intro}).
In MVPD, multi-view aggregation is essential because each view has an overlapping but different field of view.
Early methods~\cite{baque2017deep,chavdarova2017deep,fleuret2007multicamera,roig2011conditional} detected pedestrians for each view using monocular detection~\cite{ren2015faster,zhou2019objects} and performed multi-view aggregation based on the detection results in each view.
These methods optimized monocular detection and multi-view aggregation separately, resulting in suboptimal detection performances.
In recent years, end-to-end trainable methods~\cite{hou2020multiview,hou2021multiview,song2021stacked,engilberge2023two,qiu20223d,aung2024enhancing,zhang2024mahalanobis} that do not rely on monocular detection have become mainstream in MVPD.
End-to-end MVPD methods generate a BEV feature by projecting image features extracted from each view image onto a ground plane, predicting the BEV occupancy map.
The information from each view is aggregated through this BEV feature.
MVDet~\cite{hou2020multiview} is a pioneering end-to-end method that generates the BEV feature by projecting image features onto the ground plane using a perspective transformation.
To further improve detection performance, recent studies have proposed more effective projections~\cite{hou2021multiview,song2021stacked,aung2024enhancing}, data augmentations~\cite{qiu20223d,engilberge2023two,suzuki2024scene}, and network architectures~\cite{lee2023multiview,aung2024enhancing}.
Our MVPT method extends MVDet to perform the association of pedestrians utilizing information on past trajectories.

\noindent
\textbf{Pedestrian Tracking.}
Many monocular pedestrian tracking methods, including both non-deep learning~\cite{huang2012multiple,zhang2008global,perera2006multi,jiang2007linear,leibe2007coupled} and deep learning~\cite{braso2020learning,dai2021learning,zhou2022global,cetintas2023unifying,qin2023motiontrack,wang2020towards,yu2022towards,du2021giaotracker,du2023strongsort,aharon2022bot} ones, leverage information at multiple timestamps in past trajectories for robust association.
Fusion-based methods~\cite{wang2020towards,yu2022towards,du2021giaotracker,du2023strongsort,aharon2022bot} fuse the appearance features at multiple timestamps using the exponential moving average (EMA) frame-by-frame and generate the representative appearance feature for each trajectory.
Graph-based methods~\cite{braso2020learning,dai2021learning,cetintas2023unifying} construct a graph that represents the association based on appearance features at multiple timestamps and generate trajectories by optimizing the graph.
GTR~\cite{zhou2022global} extends DETR~\cite{carion2020end} to tracking and directly generates trajectories based on interactions between queries and pedestrian features across multiple timestamps.
MotionTrack~\cite{qin2023motiontrack} extracts motion information on the whole trajectory utilizing the spatial distribution pattern and velocity-time correlation based on locations over multiple past timestamps.
While these methods have improved their tracking performances on benchmark datasets~\cite{milan2016mot16,dendorfer2020mot20,sun2022dancetrack}, they cannot be applied for MVPT directly because they are designed for monocular tracking and are unable to consider the relationships between multiple views and the BEV map.
Encouraged by the success of the above monocular tracking methods, many non-end-to-end MVPT methods~\cite{berclaz2006robust,quach2021dyglip,cheng2023rest,wang2024bev,niculescu2024mctr,feng2024unveiling} also leverage information at multiple timestamps in past trajectories using techniques such as the dynamic programming, graph, and transformer~\cite{vaswani2017attention}.
However, previous end-to-end MVPT methods~\cite{teepe2024earlybird,teepe2024lifting,engilberge2023multi,yang2024end,dakic2024resource} have not incorporated this information, and it is not clear how to leverage it.
\section{Methodology}
\label{sec:methodology}

In this section, we first formulate the problem setting of our MVTrajecter.
We then provide a detailed explanation of MVTrajecter, including its overall detection and association procedure, model architecture, and training method.

\subsection{Problem Setting of MVTrajecter}
\label{subsec:problemsetting}

MVTrajecter processes a multi-view video in a sequential frame-by-frame manner, where a frame consists of images from multiple views at one timestamp.
Let $\boldsymbol{I}^t = \{ I^{t}_{1}, \cdots, I^{t}_{S} \}$ be a multi-view image at the current timestamp $t$ in a multi-view video from synchronized $S$ cameras, where $I^{t}_{s} \in \mathbb{R}^{3 \times H \times W}$ is an image from the $s$-th camera at $t$.
Here, $H$ and $W$ are the height and width of the image, respectively.
MVTrajecter aims to detect the current pedestrians from $\boldsymbol{I}^t$ and generate the current trajectories $\{{\tau}^{t-K \colon t}_{1}, \cdots, {\tau}^{t-K \colon t}_{N} \}$ of all $N$ pedestrians by associating the current pedestrian locations $\{ p^{t}_{1}, \cdots, p^{t}_{N} \}$ with past trajectories $\{ {\tau}^{t-K \colon t-1}_{1}, \cdots, {\tau}^{t-K \colon t-1}_{N} \}$, where pedestrian locations and trajectories represent those on the BEV map.
Here, $K$ is the length of past trajectories that MVTrajecter utilizes for the association.
$p^{t-k}_{n} \in \mathbb{R}^{2} \cup \{ \emptyset \}$ represents the location at timestamp $t-k$ of the pedestrian $n$, where $k=0, \cdots, K$.
$p^{t-k}_{n} = \emptyset$ indicates that the pedestrian $n$ cannot be located at $t-k$.
Trajectory ${\tau}^{t-K \colon t-k}_{n} = \{ p^{t-K}_{n}, \cdots, p^{t-k}_{n} \}$ represents a sequence of pedestrian $n$'s locations from $t-K$ to $t-k$.

We can obtain the final tracks by solving the above problem in order from the first to last timestamps in the video and linking generated trajectories pedestrian-by-pedestrian.

Note that previous end-to-end methods~\cite{teepe2024earlybird,teepe2024lifting,engilberge2023multi,yang2024end,dakic2024resource} set $K=1$ and use information on pedestrians between $t$ and $t-1$ for the association.
In contrast, MVTrajecter utilizes information on pedestrians from $t-K$ to $t$.

\begin{figure*}[tb]
    \centering
    \includegraphics[width=165mm]{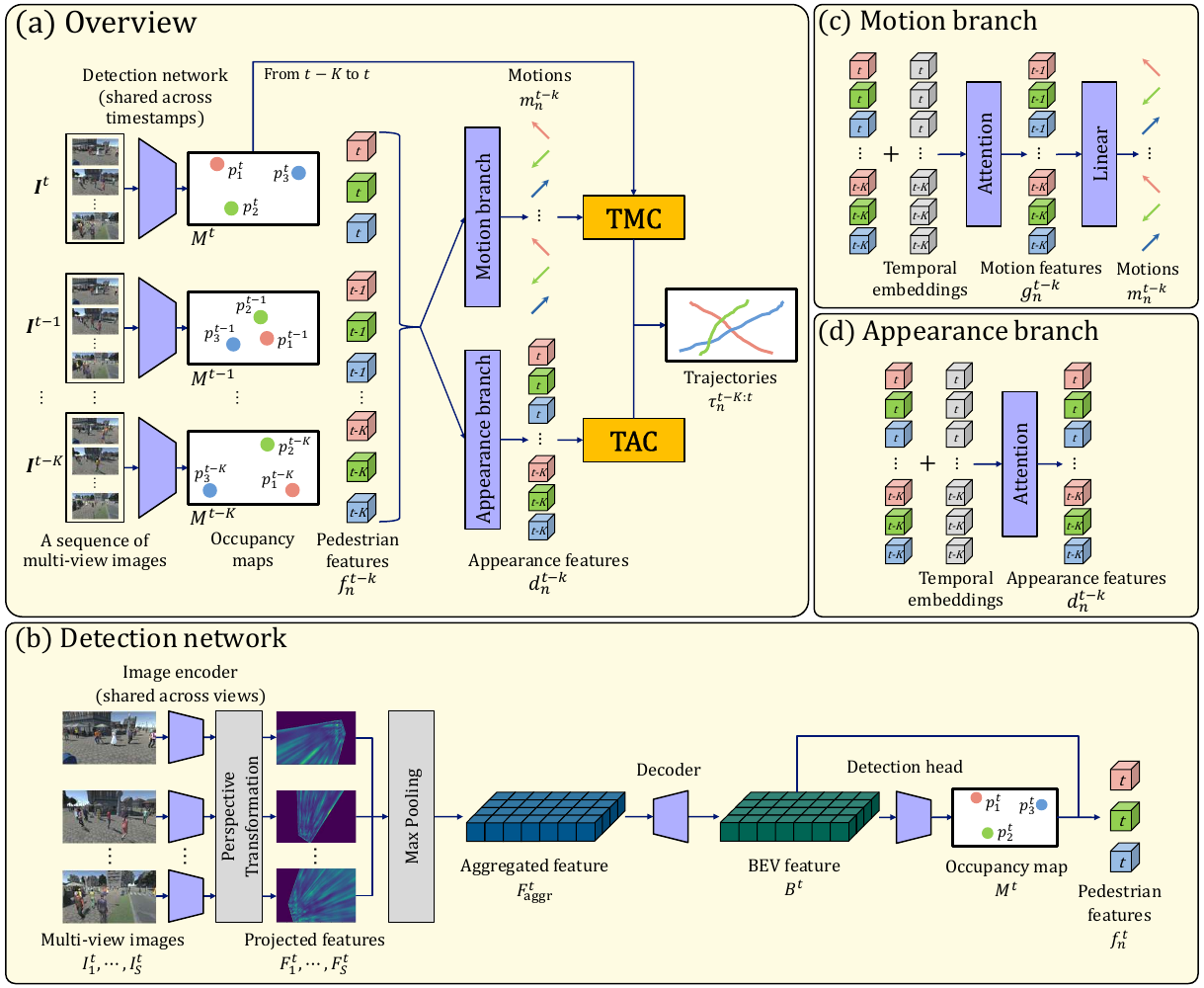}
    \vspace{-3mm}
    \caption{(a) Overview of MVTrajecter. It consists of a (b) detection network, (c) motion branch, (d) appearance branch, TMC, and TAC (see Sec.~\ref{sub:cost}). The purple blocks have trainable parameters. This is an example of the association for pedestrians at $t$.}
    \vspace{-3mm}
    \label{fig:fig2}
\end{figure*}

\subsection{Procedure of MVTrajecter}
\label{sub:procedure}

MVTrajecter consists of two parts: detecting pedestrians at the current timestamp $t$ and associating the current pedestrians with past trajectories.
In this subsection, we explain the procedures of these parts in order.

In the detection part, we predict the current BEV occupancy map $M^{t} \in \mathbb{R}^{1 \times X \times Y}$ from the current multi-view image $\boldsymbol{I}^t$, where $X$ and $Y$ are the width and height of the BEV occupancy map, respectively.
We can obtain the current pedestrian locations $\{ p^{t}_{1}, \cdots, p^{t}_{N} \}$ from $M^{t}$.

In the association part, we calculate association costs between the current pedestrian locations $\{ p^{t}_{1}, \cdots, p^{t}_{N} \}$ and past trajectories $\{ {\tau}^{t-K \colon t-1}_{1}, \cdots, {\tau}^{t-K \colon t-1}_{N} \}$, indicating which current pedestrian is likely identical to which past trajectory.
To this end, we utilize motion and appearance information of pedestrians between the current and past $K$ timestamps.
First, for each pedestrian at each past timestamp, we predict the motion to the current timestamp.
The motion to be predicted is defined as the offset between the current and past locations.
For pedestrian $n$ at $t-k$, this offset is $p^{t}_{n} - p^{t-k}_{n}$.
Then, based on the predicted motions, we calculate the motion-based association cost named trajectory motion cost (TMC).
Next, for each pedestrian at the current timestamp and past $K$ timestamps, we extract the appearance feature that represents the appearance of the pedestrian.
Then, based on the predicted appearance features, we calculate the appearance-based association cost named trajectory appearance cost (TAC).
In Sec.~\ref{sub:cost}, we will formulate TMC and TAC in detail.
After obtaining TMC and TAC, we calculate the overall association cost by fusing them to account for motion and appearance information simultaneously.
Finally, we use Hungarian Matching~\cite{kuhn1955hungarian} for the overall association cost to perform association.
As a result of these processes, we generate the current trajectories $\{ {\tau}^{t-K \colon t}_{1}, \cdots, {\tau}^{t-k \colon t}_{N} \}$.
We will explain the model architecture to implement this procedure in Sec.~\ref{sub:architecture}.

\subsection{Association Cost}
\label{sub:cost}

MVTrajecter calculates TMC and TAC and performs the association by combining them.
Let $i \in \{ 1, \cdots, N\}$ be the pedestrian whose location is $p^{t}_{i}$ at the current timestamp $t$, and $j \in \{ 1, \cdots, N\}$ be the pedestrian whose past trajectory is ${\tau}^{t-K \colon t-1}_{j}$.
We define the overall association cost between $i$ and $j$ by $C(i, j) = (1-\alpha)C_\mathrm{TMC}(i, j) + \alpha C_\mathrm{TAC}(i, j)$, where $\alpha$ is a predetermined weighting parameter.
Here, $C_\mathrm{TMC}(i, j)$ and $C_\mathrm{TAC}(i, j)$ are TMC and TAC between $i$ and $j$, respectively (we will explain how to calculate them later).
During inference, we calculate $C(i, j)$ for all pairs of $i$ and $j$ and use Hungarian Matching with a certain threshold to decide which pair of $i$ and $j$ represents identical pedestrians.
We initialized unmatched current pedestrians as new trajectories.

\noindent
\textbf{Trajectory Motion Cost (TMC).}
Let $m^{t-k}_{j} \in \mathbb{R}^2$ be the predicted motion of $j$ from timestamp $t-k$ to $t$.
When the distance $||{p}_{i}^{t} - ({p}_{j}^{t-k} + {m}_{j}^{t-k})||$ between the current detected location $(\ie, p^{t}_{i})$ and estimated current location by motion $(\ie, {p}_{j}^{t-k} + m^{t-k}_{j})$ is small, $i$ is likely identical to $j$ among pedestrians at $t-k$.
Furthermore, when this distance is large for multiple $k$, $i$ and $j$ are less likely to be identical than when the distance is large for only one $k$.
On the basis of this, we define TMC between $i$ and $j$ by summing the Euclidean distance for each past timestamp, as:
\begin{align}
    \centering
    \vspace{-10mm}
    C_\mathrm{TMC}(i, j) &= \sum_{k=1}^{K}{||p_{i}^{t} - ({p}_{j}^{t-k} + m^{t-k}_{j})||}.
\label{formula:tmc}
\end{align}
When $C_\mathrm{TMC}(i, j)$ is small, $i$ and $j$ are likely to be identical.
If $p_{j}^{t-k} = \emptyset$, we calculate $||p_{i}^{t} - ({p}_{j}^{t-k} + m^{t-k}_{j})||$ using the Kalman filter~\cite{kalman1960new}.
TMC utilizes motion information not only from an adjacent past timestamp but also from multiple timestamps in past trajectories.

\noindent
\textbf{Trajectory Appearance Cost (TAC).}
Let $d_{n}^{t-k} \in \mathbb{R}^{E}$ be the $E$-dimensional appearance feature of pedestrian $n$ at $t-k$.
When the similarity between $d_{i}^{t}$ and $d_{j}^{t-k}$ is high, $i$ is likely identical to $j$ among pedestrians at $t-k$.
Based on this, we define the conditional probability that $i$ is identical to $j$ among pedestrians at $t-k$ using the dot production as:
\begin{align}
    \centering
    \vspace{-10mm}
    \mathrm{Pr}^{t-k}(j | i) &= \frac{\exp({d_{i}^{t} \cdot d_{j}^{t-k}})}{\sum_{n=1}^{N}{\exp({d_{i}^{t} \cdot d_{n}^{t-k}}})}.
\label{formula:prob}
\end{align}
If $p_{j}^{t-k} = \emptyset$, we set $\mathrm{Pr}^{t-k}(j | i) = 0$.
When $\mathrm{Pr}^{t-k}(j | i)$ is high for multiple $k$, $i$ and $j$ are more likely to be identical than when the probability is high for only one $k$.
Therefore, we define TAC between $i$ and $j$ by summing the probability for each past timestamp, as:
\begin{align}
    \centering
    \vspace{-10mm}
    C_\mathrm{TAC}(i, j) &= - \sum_{k=1}^{K}{\mathrm{Pr}^{t-k}(j | i)}.
\label{formula:tac}
\end{align}
When $C_\mathrm{TAC}(i, j)$ is small, $i$ and $j$ are likely to be identical.
Unlike TMC, since high $\mathrm{Pr}^{t-k}(j | i)$ indicates identical pedestrian, we multiply $\mathrm{Pr}^{t-k}(j | i)$ by $-1$.
TAC utilizes appearance information not only from an adjacent past timestamp but also from multiple timestamps in past trajectories.

\subsection{Model Architecture}
\label{sub:architecture}

The overview of MVTrajecter is shown in Fig.~\ref{fig:fig2}(a).
MVTrajecter is an end-to-end trainable method that outputs everything required for our detection and association procedure.
It consists of the detection part, which predicts the BEV occupancy map and extracts pedestrian locations and pedestrian features, and the association part, which predicts motions and appearance features from pedestrian features.

\noindent
\textbf{Detection part.}
Based on~\cite{hou2020multiview,vora2023bringing,teepe2024earlybird}, the detection network consists of the image encoder, perspective transformation~\cite{hou2020multiview}, max pooling, decoder, and detection head, as shown in Fig.~\ref{fig:fig2}(b).
The input to the detection part is a multi-view image $\boldsymbol{I}^t = \{ I^{t}_{1}, \cdots, I^{t}_{S} \}$.
The image encoder extracts $E$-channel feature maps from each view image, and then, the extracted feature maps are projected onto a ground plane by perspective transformation~\cite{hou2020multiview} (formulated in Supp.~\ref{sup:perspective}).
We denote these projected features as $\{ F^t_1, \cdots, F^t_S \}$, where $F^t_s \in \mathbb{R}^{E \times X \times Y}$ is the projected feature of the $s$-th camera at $t$.
The max pooling operation aggregates projected features along the camera direction as:
\begin{align}
    \centering
    \vspace{-10mm}
    &F^t_{\mathrm{aggr}} (e,x,y)=  {\underset{s \in  \{ 1, \cdots, S\}}{\mathrm{max}}} F^t_s(e,x,y),
\label{formula:aggr}
\end{align}
where $F^t_{\mathrm{aggr}}$ is referred to as the aggregated feature, and $F^t_s(e,x,y)$ is the value at coordinates $(e,x,y)$ on $F^t_s$.
This aggregation allows the later modules to handle any number of cameras and camera layouts~\cite{vora2023bringing}.
The decoder refines $F^t_{\mathrm{aggr}}$ and generates a BEV feature $B^t \in \mathbb{R}^{E \times X \times Y}$.
The detection head predicts a BEV occupancy map $M^t \in \mathbb{R}^{1 \times X \times Y}$ from $B^t$.
We apply a $3 \times 3$ max pooling to $M^t$ following~\cite{zhou2019objects} and extract pedestrians over a certain threshold.
As a result, we obtain current pedestrian locations $\{ p^{t}_{1}, \cdots, p^{t}_{N} \}$.
We also extract pedestrian features by $f_{n}^{t} = B^{t} (p_{n}^{t})$ for the subsequent association part, where $f_{n}^{t}$ is a pedestrian feature of pedestrian $n$ at $t$, and $B^{t} (x,y) \in \mathbb{R}^{E}$ represents a feature vector at coordinates $(x, y)$ on $B^{t}$.

\noindent
\textbf{Association part.}
The association part predicts motions and appearance features from current and past pedestrian features $\{ f_{1}^{t-k}, \cdots, f_{N}^{t-k} \}_{k=0}^{K}$.
We keep past pedestrian features $\{ f_{1}^{t-k}, \cdots, f_{N}^{t-k} \}_{k=1}^{K}$ and do not need to re-calculate them.
The association part has a motion branch to predict motions and an appearance branch to predict appearance features.
We utilize the attention mechanism~\cite{vaswani2017attention} for both branches.
The attention mechanism has the ability to capture the multiple temporal changes of pedestrians~\cite{zeng2022motr,meinhardt2022trackformer,xu2022transcenter} and match pedestrians with high similarity at different timestamps~\cite{fukui2023multiobject,zhu2018online,zhou2022global}.
The former ability fits to predict motions over multiple timestamps and the latter ability fits to predict identifiable appearance features.
Before feeding pedestrian features into the attention mechanism, we add sinusoidal temporal embeddings~\cite{vaswani2017attention} to them in order to distinguish the timestamps (from $t-K$ to $t$) of pedestrian features.

The motion branch consists of an attention layer and a linear layer, as shown in Fig.~\ref{fig:fig2}(c).
The attention layer captures the temporal changes of pedestrians from pedestrian features, generating motion features $\{ {g}_{1}^{t-k}, \cdots, {g}_{N}^{t-k} \}_{k=1}^{K}$.
Then, the linear layer predicts motions $\{ m_{1}^{t-k}, \cdots, m_{N}^{t-k} \}_{k=1}^{K}$ from motion features:
\begin{align}
    \centering
    \vspace{-10mm}
    &{g}_{n}^{t-k} = \mathrm{Attention}(f_{n}^{t-k}, \{ f_{1}^{t-k}, \cdots, f_{N}^{t-k} \}_{k=0}^{K} ), \\
    &m_{n}^{t-k} = \mathrm{Linear} (g_{n}^{t-k}),
\label{formula:motion_attn}
\end{align}
where $g^{t-k}_{n} \in \mathbb{R}^{E}$ is the motion feature of pedestrian $n$ at $t-k$, and $m^{t-k}_{n} \in \mathbb{R}^2$ is the motion of pedestrian $n$ from $t-k$ to $t$.
For the $\mathrm{Attention}$ function, the first argument indicates query, and the second argument indicates key and value.
These predicted motions are utilized for TMC.

The appearance branch consists of an attention layer, as shown in Fig.~\ref{fig:fig2}(d).
This generates appearance features $\{ d_{1}^{t-k}, \cdots, d_{N}^{t-k} \}_{k=0}^{K}$ by increasing the similarity between identical pedestrian's features at different timestamps, as:
\begin{align}
    \centering
    \vspace{-10mm}
    &d_{n}^{t-k} =  \mathrm{Attention}(f_{n}^{t-k}, \{ f_{1}^{t-k}, \cdots, f_{N}^{t-k} \}_{k=0}^{K}),
\label{formula:appearance_attn}
\end{align}
where $d^{t-k}_{n} \in \mathbb{R}^E$ is the appearance feature of pedestrian $n$ at $t-k$.
These appearance features are utilized for TAC.

\subsection{Training}
\label{sub:train}

During training, we split multi-view videos into non-overlapping multi-view image sequences with window size $K+1$ and feed $\{ \boldsymbol{I}^{t-K}, \cdots, \boldsymbol{I}^{t} \}$ into MVTrajecter.
We train the entire MVTrajecter in an end-to-end manner.
The overall training objective consists of detection loss, the loss related to TMC, and the loss related to TAC.
First, we predict occupancy maps $\{ M^{t-K}, \cdots, M^{t} \}$ from $\{ \boldsymbol{I}^{t-K}, \cdots, \boldsymbol{I}^{t} \}$.
The detection loss $\mathcal{L}_\mathrm{det}$ is calculated using a mean squared error (MSE) for all predicted occupancy maps $\{ M^{t-K}, \cdots, M^{t} \}$, as:
\begin{align}
    \centering
    \vspace{-10mm}
    &\mathcal{L}_\mathrm{det} = \frac{1}{K+1} \sum_{k=0}^{K}\mathrm{MSE}(M^{t-k}, \bar{M}^{t-k}),
\label{formula:loss_det}
\end{align}
where $\bar{M}^{t-k}$ is a smoothed ground truth occupancy map at $t-k$ using a Gaussian kernel following~\cite{hou2020multiview}.

Then, we predict motions $\{ m^{t-k}_{1}, \cdots, m^{t-k}_{N} \}_{k=1}^{K}$ and appearance features $\{ d^{t-k}_{1}, \cdots, d^{t-k}_{N} \}_{k=0}^{K}$.
Let $\mathcal{L}_\mathrm{TMC}$ be the loss related to TMC.
For pedestrian $n$ and $k=1, \cdots,K$, we minimize $||p_{n}^{t} - ({p}_{n}^{t-k} + m_{n}^{t-k})||$.
We optimize $\{ m_{n}^{t-K}, \cdots, m_{n}^{t-1} \}$ because $\{ p_{n}^{t-K}, \cdots, p_{n}^{t} \}$ are fixed by the detection part.
Therefore, we define $\mathcal{L}_\mathrm{TMC}$ as:
\begin{align}
    \centering
    \vspace{-10mm}
    &\mathcal{L}_\mathrm{TMC} = \frac{1}{N} \frac{1}{K} \sum_{n=1}^{N} \sum_{k=1}^{K} ||m_{n}^{t-k} - (\bar{p}^{t}_{n} - \bar{p}^{t-k}_{n})||,
\label{formula:loss_tmc}
\end{align}
where $\bar{p}^{t-k}_{n}$ represents the ground truth location of pedestrian $n$ at $t-k$.
Let $\mathcal{L}_\mathrm{TAC}$ be the loss related to TAC.
For pedestrian $n$ and $k=1, \cdots,K$, we maximize $\mathrm{Pr}^{t-k}(j=n | i=n)$.
Therefore, we define $\mathcal{L}_\mathrm{TAC}$ using negative log-likelihood, as:
\begin{align}
    \centering
    \vspace{-10mm}
    &\mathcal{L}_\mathrm{TAC} = - \frac{1}{N} \frac{1}{K} \sum_{n=1}^{N} \sum_{k=1}^{K} \log  \mathrm{Pr}^{t-k}(j = n | i = n).
\label{formula:loss_tac}
\end{align}

We set the overall loss $\mathcal{L}_\mathrm{all}$ using the uncertainty loss~\cite{kendall2018multi,zhang2021fairmot} to balance $\mathcal{L}_\mathrm{det}$, $\mathcal{L}_\mathrm{TMC}$, and $\mathcal{L}_\mathrm{TAC}$ as follows:
\begin{align}
    \centering
    \vspace{-10mm}
    \mathcal{L}_\mathrm{all} = \frac{1}{e^{w_1}}\mathcal{L}_\mathrm{det} + \frac{1}{e^{w_2}}\mathcal{L}_\mathrm{TMC} &+ \frac{1}{e^{w_3}}\mathcal{L}_\mathrm{TAC} \notag \\
    &+ w_1 + w_2 + w_3,
\label{formula:loss}
\end{align}
where $w_1$, $w_2$, and $w_3$ are learnable parameters that automatically balance the three losses.
\section{Experiments}
\label{sec:experiment}

\subsection{Datasets}
\label{sub:data}

In this subsection, we explain the three MVPT datasets used in our experiments.
More details can be found in Supp.~\ref{sup:data_detail}.

\noindent
\textbf{Wildtrack}~\cite{chavdarova2018wildtrack}.
This is a real-world dataset consisting of videos captured with $7$ cameras.
The videos comprise $400$ frames at a frame rate of $2$ fps.
This dataset covers a $12 \, \mathrm{m} \times 36 \, \mathrm{m}$ region quantized into a $480 \times 1440$ grid using square grid cells of $2.5 \, \mathrm{cm^2}$.
Each frame includes $20$ pedestrians on average.
Wildtrack splits them into the first $360$ frames for training and the remaining $40$ frames for testing.
We use the last $40$ frames in the training split as the validation data.

\noindent
\textbf{MultiviewX}~\cite{hou2020multiview}.
This is a synthetic dataset using the Unity engine and closely follows the style of Wildtrack.
This dataset consists of videos captured by $6$ cameras and covers a $16 \, \mathrm{m} \times 25 \, \mathrm{m}$ region.
The region is quantized into a $640 \times 1000$ grid.
Each frame includes $40$ pedestrians on average.
MultiviewX also splits them into the first $360$ frames for training and the remaining $40$ frames for testing.
We use the last $40$ frames in the training split as the validation data.

\noindent
\textbf{GMVD}~\cite{vora2023bringing}.
This is a large-scale synthetic dataset and includes $7$ scenes with varying numbers of cameras and camera layouts.
Furthermore, each scene contains multiple sequences with various environmental conditions including time and weather.
This makes GMVD more challenging than Wildtrack or MultiviewX.
Except for the camera setup and the size of the covered region, all parameters follow those of MultiviewX.
Each frame includes $20\text{--}40$ pedestrians on average.
GMVD splits them into $6$ scenes (with $43$ sequences and $4983$ frames) for training and $1$ scene (with $10$ sequences and $1012$ frames) for testing.
Following Vora~et al.~\cite{vora2023bringing}, we use MultiviewX as the validation data in the experiments on GMVD.
Since GMVD is the largest dataset, we set it as the main dataset for our experiments.

\subsection{Evaluation Metrics}
\label{sub:set}

Following previous studies~\cite{cheng2023rest,teepe2024earlybird,teepe2024lifting}, we used five standard metrics provided by Kasturi~et al.~\cite{kasturi2008framework} and Ristan~et al.\cite{ristani2016performance}: IDF1, Multiple Object Tracking Accuracy (MOTA), Multiple Object Tracking Precision (MOTP), Mostly Tracked (MT), and Mostly Lost (ML).
A predicted pedestrian was classified as a true positive if its distance from the ground truth was within $1.0$ meters.
We used IDF1 and MOTA as the primary performance indicators following previous studies~\cite{teepe2024earlybird,teepe2024lifting,yang2024end}.

\subsection{Implementation Details}
\label{sub:implementation}

Following previous methods~\cite{hou2020multiview,teepe2024earlybird,teepe2024lifting}, we employed ResNet18~\cite{he2016deep}, ResNet18 U-Net~\cite{ronneberger2015u}, and a $2$-layer convolutional neural network for the image encoder, decoder, and detection head in the detection network, respectively.
ResNet18 was pretrained on ImageNet~\cite{deng2009imagenet}.
We employed $2$ transformer attention blocks~\cite{vaswani2017attention} with $8$ attention heads for each attention layer.

The input images were resized to $720 \times 1280$.
We set the channel size $E$ to $1024$.
Unless otherwise mentioned, we set the temporal length $K$ of past trajectories to $7$, which is the maximum size we can set due to the limitations of GPU memory.
We set the weighting parameter $\alpha$ between TMC and TAC to $0.98$, the threshold of detection to $0.4$, and the threshold of Hungarian Matching to $0.1$.
These hyperparameters were tuned on the validation data.

We optimized MVTrajecter using the Adam optimizer~\cite{loshchilov2018decoupled}.
We set the batch size to $1$ and accumulated the gradient over $16$ batches.
The learning rate was initialized to $1.0 \times 10^{-3}$ and was decayed to $1.0 \times 10^{-6}$ following a cosine schedule.
We set the maximum training epochs to $20$ for GMVD and $50$ for Wildtrack and MultiviewX and stopped the training if the overall loss for the validation data did not decrease for $3$ epochs in a row.
More detailed implementations are provided in Supp.~\ref{sup:implementation}.
All experiments were conducted on a Nvidia 80GB A100 GPU.

\begin{table}[t]
\centering
\vspace{-2.0mm}
\scalebox{0.82}{
\begin{tabular}{c|ccccc} \hline
Method & IDF1$\uparrow$ & MOTA$\uparrow$ & MOTP$\uparrow$  & MT$\uparrow$ & ML$\downarrow$ \\ \hline
EarlyBird~\cite{teepe2024earlybird} & 69.7 & 70.8 & 86.1 & 41.6 & 12.4 \\ 
MVFlow~\cite{engilberge2023multi} & 70.5 & 71.1 & 85.7 & 41.0 & 12.7 \\ 
TrackTacular~\cite{teepe2024lifting} & 74.4 & 71.2 & 85.2 & 52.4 & 12.5 \\
MVTr~\cite{yang2024end} & 74.3 & 75.6 & 84.2 & 63.1 & 10.3 \\
Ours & \textbf{77.2} & \textbf{78.1} & \textbf{87.7} & \textbf{66.0} & \textbf{8.7} \\ \hline
\end{tabular}
}
\vspace{-2.0mm}
\caption{Comparison with previous methods on GMVD when all models used their own detection results for tracking.}
\vspace{-2.0mm}
\label{table:gmvd_det_track}
\end{table}

\begin{table}[t]
\centering
\vspace{-1.0mm}
\scalebox{0.82}{
\begin{tabular}{c|ccccc} \hline
Method & IDF1$\uparrow$ & MOTA$\uparrow$ & MOTP$\uparrow$ & MT$\uparrow$ & ML$\downarrow$ \\ \hline
EarlyBird~\cite{teepe2024earlybird} & 70.6 & 72.6 & 85.9 & 43.9 & 11.6 \\ 
MVFlow~\cite{engilberge2023multi} & 73.2 & 73.9 & 83.7 & 58.1  & 9.0 \\ 
TrackTacular~\cite{teepe2024lifting} & 75.6 & 76.2 & 84.9 & 64.3 & 8.8 \\ 
MVTr~\cite{yang2024end} & 75.3 & 76.0 & 85.0 & 64.1 & 9.0 \\
Ours & \textbf{77.2} & \textbf{78.1} & \textbf{87.7} & \textbf{66.0} & \textbf{8.7} \\ \hline
\end{tabular}
}
\vspace{-2.0mm}
\caption{Comparison with previous methods on GMVD when all models used the same detection results as ours for tracking.}
\vspace{-2.0mm}
\label{table:gmvd_track}
\end{table}

\subsection{Comparison with State-of-the-Art Methods}
\label{sub:result}

To verify the effectiveness of MVTrajecter, we compared it with previous state-of-the-art methods.
Table~\ref{table:gmvd_det_track} shows the comparison results on GMVD.
Each method used its own detection results for tracking.
Since the previous methods had not been evaluated on GMVD, we re-implemented EarlyBird~\cite{teepe2024earlybird}, MVFlow~\cite{engilberge2023multi}, TrackTacular~\cite{teepe2024lifting}, and MVTr~\cite{yang2024end}, which are the most recent state-of-the-art methods.
Their network architectures and training settings followed our implementation details as much as possible.
MVTrajecter outperformed the previous methods on all metrics.
These results demonstrate that MVTrajecter is superior to the other methods in terms of end-to-end tracking performance, including detection and association.

Since tracking results are affected by detection results, we also compared the tracking performance using the same detection results to validate the effectiveness of MVTrajecter's association only.
Table~\ref{table:gmvd_track} shows the comparison results on GMVD using the same detection results as MVTrajecter.
MVTrajecter outperformed all the comparison methods.
This demonstrates that MVTrajecter's association, which uses motion and appearance information over multiple past timestamps, is superior to the association of the other methods.
Specifically, its improvement over EarlyBird, which mainly depends on appearance information for the association, is due to utilizing information from multiple past timestamps and incorporating motion information.
Improvements over MVFlow, TrackTacular, and MVTr, which only use motion information for the association, are due to utilizing information from multiple past timestamps and incorporating appearance information. 
Furthermore, all comparison methods improved their tracking performances when using MVTrajecter's detection results.
This indicates that MVTrajecter's detection is superior to those of the other methods.
A detailed evaluation and analysis of MVTrajecter's detection are provided in Supp.~\ref{sup:detection}.

To validate the versatility of MVTrajecter, which is not limited to GMVD, we also experimented using MultiviewX and Wildtrack.
Tables~\ref{table:wildtrack} and \ref{table:multiviewx} show the comparison results on Wildtrack and MultiviewX, respectively, where the results of the previous methods are directly copied from the original papers.
MVTrajecter outperformed the previous methods on both Wildtrack and MultiviewX, achieving new state-of-the-art tracking performances of $94.3$ MOTA on Wildtrack and $92.8$ MOTA on MultiviewX.
These results demonstrate that MVTrajecter is effective for various datasets.
Qualitative results are provided in Supp.~\ref{sup:qualiative}.

\begin{table}[t]
\centering
\vspace{-2.0mm}
\scalebox{0.82}{
\begin{tabular}{c|ccccc} \hline
Method & IDF1$\uparrow$ & MOTA$\uparrow$ & MOTP$\uparrow$ & MT$\uparrow$ & ML$\downarrow$ \\ \hline
$\textrm{ReST}^{\dagger}$~\cite{cheng2023rest} & 85.7 & 81.6 & 81.8 & 79.4 & 4.9 \\
$\textrm{BEV-SUSHI}^{\dagger}$~\cite{wang2024bev} & 93.4 & 87.5 & \textbf{94.3} & 90.2 & \textbf{2.4} \\
$\textrm{REMP}^{\dagger}$~\cite{dakic2024resource} & -- & 88.5 & 86.8 & -- & --  \\
EarlyBird~\cite{teepe2024earlybird} & 92.3 & 89.5 & 86.6 & 78.0 & 4.9 \\ 
MVFlow~\cite{engilberge2023multi} & 93.5 & 91.3 & -- & -- & -- \\ 
TrackTacular~\cite{teepe2024lifting} & 95.3 & 91.8 & 85.4 & 87.8 & 4.9 \\ 
MVTr~\cite{yang2024end} & 93.1 & 92.3 & 92.7 & \textbf{95.1} & 4.9 \\
Ours & \textbf{96.5} & \textbf{94.3} & 93.0 & 90.2 & 4.9 \\ \hline
\end{tabular}
}
\vspace{-2.0mm}
\caption{Comparison with previous methods on Wildtrack. Methods with $\dagger$ used external data or external models.
}
\vspace{-2.0mm}
\label{table:wildtrack}
\end{table}

\begin{table}[t]
\centering
\vspace{-1.0mm}
\scalebox{0.82}{
\begin{tabular}{c|ccccc} \hline
Method & IDF1$\uparrow$ & MOTA$\uparrow$ & MOTP$\uparrow$ & MT$\uparrow$ & ML$\downarrow$ \\ \hline
$\textrm{REMP}^{\dagger}$~\cite{dakic2024resource} & -- & 81.0 & 85.8 & -- & --  \\
EarlyBird~\cite{teepe2024earlybird} & 82.4 & 88.4 & 86.2 & 82.9 & 1.3 \\ 
TrackTacular~\cite{teepe2024lifting} & 85.6 & 92.4 & 80.1 & 92.1 & 2.6 \\
MVTr~\cite{yang2024end} & 82.9 & 91.4 & \textbf{95.0} & 96.1 & \textbf{0.0} \\
Ours & \textbf{85.8} & \textbf{92.8} & \textbf{95.0} & \textbf{97.4} & \textbf{0.0} \\ \hline
\end{tabular}
}
\vspace{-2.0mm}
\caption{Comparison with previous methods on MultiviewX. The method with $\dagger$ used external data or external models.
}
\vspace{-2.0mm}
\label{table:multiviewx}
\end{table}

\subsection{Ablation Study}
\label{sub:ablation}

We conducted ablation studies to investigate the effect of past trajectory length, TMC, and TAC, as well as the advantages of employing the attention mechanism to predict motions and appearance features.
All experiments were conducted on GMVD.
We also conducted more detailed ablation studies in Supp.~\ref{sup:ablation} and analyzed the inference speed of our proposed method in Supp.~\ref{sup:speed}.

\noindent
\textbf{Effect of Past Trajectory Length $K$}.
Table~\ref{table:length} shows the comparison results using different lengths $K$ of past trajectories.
When $K=1$, our method is similar to the comparison methods in Table~\ref{table:gmvd_det_track} and discards the information on past trajectories before an adjacent past timestamp, yielding a low tracking performance.
As we increased $K$, we consistently observed improvements in MOTA and IDF1, and our method outperformed the comparison methods in Table~\ref{table:gmvd_det_track} even when $K=5$.
The largest $K=7$ achieved $3.9$ points higher MOTA and $6.2$ points higher IDF1 than the smallest $K=1$.
These performance improvements demonstrate the advantage of using longer trajectories for the association.

\noindent
\textbf{Effect of TMC and TAC}.
To investigate the effect of TMC and TAC on the tracking performance, we implemented two baselines of MVTrajecter, an appearance baseline and a motion baseline, and added TMC and TAC to them. 
The motion and appearance baselines only utilize the motion and appearance information between the current and an adjacent past timestamp, respectively.
Table~\ref{table:tmc_tac} shows the impacts of TMC and TAC on tracking performances.
When TMC was added to the motion baseline or TAC was added to the appearance baseline, their tracking performances were greatly improved.
These results demonstrate the effectiveness of utilizing the information for multiple past timestamps via TMC and TAC.
In addition, when both TMC and TAC were added to the baselines, tracking performances were improved further.
This indicates that both motion and appearance information from multiple past timestamps are necessary for a better tracking performance.

\begin{table}[t]
\centering
\vspace{-2.0mm}
\scalebox{0.90}{
\begin{tabular}{c|ccccc} \hline
$K$ & IDF1$\uparrow$ & MOTA$\uparrow$ & MOTP$\uparrow$ & MT$\uparrow$ & ML$\downarrow$ \\ \hline
$1$ & 71.0 & 74.2 & 87.0 & 60.7 & 9.0 \\ 
$3$ & 73.6 & 76.4 & 87.3 & 63.5 & 8.8 \\ 
$5$ & 76.0 & 77.2 & \textbf{87.8} & 64.8 & \textbf{8.7} \\ 
$7$ & \textbf{77.2} & \textbf{78.1} & 87.7 & \textbf{66.0} & \textbf{8.7} \\ \hline
\end{tabular}
}
\vspace{-2.0mm}
\caption{Effect of past trajectory length $K$.
}
\vspace{-2.0mm}
\label{table:length}
\end{table}

\begin{table}[t]
\centering
\vspace{-1.0mm}
\scalebox{0.80}{
\begin{tabular}{c|ccccc} \hline
& IDF1$\uparrow$ & MOTA$\uparrow$ & MOTP$\uparrow$ & MT$\uparrow$ & ML$\downarrow$ \\ \hline
Motion baseline & 67.8 & 73.7 & 86.2 & 58.1 & 9.0 \\ 
+ TMC & 71.8 & 74.2 & 86.6 & 61.9 & \textbf{8.7} \\ 
Appearance baseline & 64.7 & 72.3 & 87.0 & 58.4 & 9.0 \\
+ TAC & 72.3 & 73.8 & 87.3 & 62.3 & 8.8 \\
+ TMC and TAC & \textbf{77.2} & \textbf{78.1} & \textbf{87.7} & \textbf{66.0} & \textbf{8.7} \\ \hline
\end{tabular}
}
\vspace{-2.0mm}
\caption{Effect of TMC and TAC.
}
\vspace{-2.0mm}
\label{table:tmc_tac}
\end{table}

\noindent
\textbf{Effectiveness of Employing the Attention Mechanism to Predict Motions.}
While we employed the attention mechanism to predict motions, we could also employ the Kalman filter~\cite{kalman1960new} which is often used in monocular tracking to predict motions.
We compared the attention mechanism with the Kalman filter to verify the advantage of employing the attention mechanism.
Table~\ref{table:motion} shows the comparison results.
Employing the attention mechanism achieved a better tracking performance on almost all metrics than employing the Kalman filter.
This result demonstrates that the motions predicted by the attention mechanism are more accurate than those predicted by the Kalman filter. 
We presume this is because the attention mechanism captures the relationships between multiple timestamps more accurately.

\noindent
\textbf{Effectiveness of Employing the Attention Mechanism to Extract Appearance Features.}
We re-extracted appearance features for all timestamps within the time window (\ie, from $t-K$ to $t$) as $t$ changed by employing the attention mechanism.
In contrast, EarlyBird~\cite{teepe2024earlybird} and REMP~\cite{dakic2024resource} extract appearance features only for $t$.
To investigate the advantage of our appearance feature extraction, we compared it with EarlyBird's appearance feature extraction.
Table~\ref{table:appearance} shows the comparison results.
Our appearance feature extraction achieved a higher tracking performance than EarlyBird's.
We presume this is because as $t$ changes, important appearance information also changes, and the attention mechanism has the ability to adaptively select that information based on the relationships among multiple timestamps.

\begin{table}[t]
\centering
\vspace{-2.0mm}
\scalebox{0.85}{
\begin{tabular}{c|ccccc} \hline
& IDF1$\uparrow$ & MOTA$\uparrow$ & MOTP$\uparrow$ & MT$\uparrow$ & ML$\downarrow$ \\ \hline 
Ours w/ KF & 76.0 & 75.3 & \textbf{87.8} & 64.1 & 8.8 \\  
Ours & \textbf{77.2} & \textbf{78.1} & 87.7 & \textbf{66.0} & \textbf{8.7} \\ \hline
\end{tabular}
}
\vspace{-2.0mm}
\caption{Comparison of motion prediction. "w/ KF" predicts motions using the Kalman filter~\cite{kalman1960new}.}
\vspace{-2.0mm}
\label{table:motion}
\end{table}

\begin{table}[t]
\centering
\vspace{-1.0mm}
\scalebox{0.85}{
\begin{tabular}{c|ccccc} \hline
& IDF1$\uparrow$ & MOTA$\uparrow$ & MOTP$\uparrow$ & MT$\uparrow$ & ML$\downarrow$ \\ \hline
Ours w/ EB & 75.6 & 76.2 & 87.4 & 62.9 & 8.8 \\  
Ours & \textbf{77.2} & \textbf{78.1} & \textbf{87.7} & \textbf{66.0} & \textbf{8.7} \\ \hline
\end{tabular}
}
\vspace{-2.0mm}
\caption{Comparison of appearance feature extraction. "w/ EB" extracts appearance features in the same way as EarlyBird~\cite{teepe2024earlybird}.}
\vspace{-2.0mm}
\label{table:appearance}
\end{table}

\section{Conclusion}
\label{sec:conclusion}

We proposed a new end-to-end multi-view pedestrian tracking~(MVPT) method called MVTrajecter.
It comprehensively identifies which current pedestrian is identical to which past trajectory based on the motion and appearance information between the current and multiple timestamps in past trajectories.
In addition, it effectively captures the relationships between multiple timestamps employing the attention mechanism.
We demonstrated the effectiveness of each component in MVTrajecter, and MVTrajecter outperformed previous methods on three major MVPT datasets.

Even though we found that utilizing information from multiple timestamps in past trajectories leads to a better tracking performance, we only utilized information at the past $7$ timestamps due to the limitations of GPU memory.
Therefore, MVTrajecter cannot track someone whose detection continuously fails in over $7$ timestamps.
In future work, we plan to explore efficient ways to utilize information over the longer term.
{
    \small
    \bibliographystyle{ieeenat_fullname}
    \bibliography{main}
}

\clearpage
\setcounter{page}{1}
\maketitlesupplementary

\begin{table*}[t]
\centering
\scalebox{1.0}{
\begin{tabular}{ccccccccc} \hline
Dataset & Scenes & Sequences & Frames & Cameras & Covered region & Grid size & Pedestrians \\ \hline \hline
Wildtrack~\cite{chavdarova2018wildtrack} & Real & 1 & 400 & 7 & $12 \times 36 \, \mathrm{m}$  & $480 \times 1440$ & 20 / frame \\ \hline 
MultiviewX~\cite{hou2020multiview} & Unity & 1 & 400 & 6 & $16 \times 25 \, \mathrm{m}$  & $640 \times 1000$ & 40 / frame \\ \hline
\multirow{7}{*}{GMVD~\cite{vora2023bringing}} & Unity & 2 & 723 & 6 & $16 \times 25 \, \mathrm{m}$  & $640 \times 1000$ & 40 / frame \\ 
& GTA & 10 & 1034 & 5 & $20 \times 30 \, \mathrm{m}$  & $800 \times 1200$ & 20  / frame \\ 
& GTA & 10 & 1000 & 3 & $30 \times 12 \, \mathrm{m}$  & $1200 \times 480$ & 30 / frame \\ 
& GTA & 10 & 1014 & 5 & $25 \times 25 \, \mathrm{m}$  & $1000 \times 1000$ & 30 / frame \\ 
& GTA & 1 & 182 & 5 & $28 \times 27 \, \mathrm{m}$  & $1120 \times 1080$ & 20 / frame \\ 
& GTA & 10 & 1030 & 7 & $33 \times 31 \, \mathrm{m}$  & $1320 \times 1240$ & 30 / frame \\ 
& GTA & 10 & 1012 & 6, 8 & $29 \times 19 \, \mathrm{m}$  & $1160 \times 760$ & 30 / frame \\ \hline 
\end{tabular}
}
\caption{Statistics of Wildtrack, MultiviewX, and GMVD. The scene in the bottom row of GMVD contains sequences of the same scene captured by $6$ or $8$ cameras. In the experiments on GMVD, we utilized the scene in the bottom row for testing and the other scene for training.}
\label{table:data_statistics}
\end{table*}

\appendix

\section{Perspective Transformation}
\label{sup:perspective}
In this section, we formulate the perspective transformation~\cite{hou2020multiview} that we used in the detection network of MVTrajecter.
As described in Sec.~\ref{sub:architecture}, the perspective transformation projects image feature maps of each view into a ground plane.
This perspective transformation is defined using 3D locations $(x,y,z)$ and 2D image pixel coordinates $(u,v)$, as:
\begin{align}
    \centering
    \vspace{-10mm}
    \gamma \begin{pmatrix}
        u \\
        v \\
        1
    \end{pmatrix}  &=  P_{\theta} \begin{pmatrix}
        x \\
        y \\
        z \\
        1
    \end{pmatrix} = A \lbrack R | T \rbrack \begin{pmatrix}
        x \\
        y \\
        z \\
        1
    \end{pmatrix} \notag \\
    &= \begin{pmatrix}
        \theta_{11} & \theta_{12} & \theta_{13} & \theta_{14} \\
        \theta_{21} & \theta_{22} & \theta_{23} & \theta_{24} \\
        \theta_{31} & \theta_{32} & \theta_{33} & \theta_{34} 
    \end{pmatrix} \begin{pmatrix}
        x \\
        y \\
        z \\
        1
    \end{pmatrix},
\label{formula:perspective}
\end{align}
where $\gamma$ is a real-valued scaling factor, and $P_{\theta}$ is the $3 \times 4$ transformation matrix calculated using the $3 \times 3$ intrinsic camera parameter matrix $A$ and the $3 \times 4$ extrinsic camera parameter matrix $\lbrack R | T \rbrack$.
Specifically, $R$ represents the rotation, and $T$ represents the translation.
By setting $z=0$, we can retrieve the correspondence between the image pixel $(u,v)$ and the ground plane coordinates $(x,y)$, as:
\begin{align}
    \centering
    \vspace{-10mm}
    \gamma \begin{pmatrix}
        u \\
        v \\
        1
    \end{pmatrix}  =  P_{\theta, 0} \begin{pmatrix}
        x \\
        y \\
        1
    \end{pmatrix} = \begin{pmatrix}
        \theta_{11} & \theta_{12} & \theta_{14} \\
        \theta_{21} & \theta_{22} & \theta_{24} \\
        \theta_{31} & \theta_{32} & \theta_{34} 
    \end{pmatrix} \begin{pmatrix}
        x \\
        y \\
        1
    \end{pmatrix},
\label{formula:perspective_0}
\end{align}
where $P_{\theta, 0}$ is the $3 \times 3$ transformation matrix that is $P_{\theta}$ with the third column canceled.
This allows the projection of image features onto the ground plane.

\section{Datasets Details}
\label{sup:data_detail}

In this section, we describe the datasets that we used in the experiments in detail.
The statistics of each dataset are summarized in Table~\ref {table:data_statistics}, and sample frames of each dataset are provided in Fig.~\ref{fig:dataset}.

\paragraph{Wildtrack~\cite{chavdarova2018wildtrack}}
This is a real-world multi-view dataset consisting of one multi-view video sequence captured with $7$ cameras.
The multi-view video comprises $400$ frames at a frame rate of $2$ fps and covers a total of $200$ seconds.
Each view video is recorded at $1080 \times 1920$ resolution.
A total of $313$ pedestrians are contained in the multi-view video, and $20$ pedestrians are contained in each frame on average.
This dataset covers a $12 \times 36 \, \mathrm{m}$ region quantized into a $480 \times 1440$ grid using square grid cells of $2.5 \, \mathrm{cm^2}$.
Each grid cell is captured by $3.74$ cameras on average.
Wildtrack splits them into the first $360$ frames ($180$ seconds) for training and the remaining $40$ frames ($20$ seconds) for testing.

\paragraph{MultiviewX~\cite{hou2020multiview}}
This is a synthetic dataset created using the Unity engine and captures a more crowded scene than Wildtrack.
MultiviewX consists of one multi-view video sequence captured with $6$ cameras.
The multi-view video comprises $400$ frames at a frame rate of $2$ fps and covers a total of $200$ seconds.
Each view video is recorded at $1080 \times 1920$ resolution.
A total of $350$ pedestrians are contained in the multi-view video, and $40$ pedestrians are contained in each frame on average.
This dataset covers a $16 \times 25 \, \mathrm{m}$ region quantized into a $640 \times 1000$ grid using square grid cells of $2.5 \, \mathrm{cm^2}$, which is slightly smaller than Wildtrack.
Each grid cell is captured by $4.41$ cameras on average.
MultiviewX also splits them into the first $360$ frames ($180$ seconds) for training and the remaining $40$ frames ($20$ seconds) for testing.

\paragraph{GMVD~\cite{vora2023bringing}}
This is a large-scale synthetic dataset that includes $7$ scenes with varying numbers of cameras and camera layouts.
Of the $7$ scenes, $6$ scenes are captured using Grand Theft Auto~(GTA), and the remaining $1$ scene using the Unity engine.
In addition, each scene also contains multiple multi-view video sequences with different environmental conditions including time and weather.
In total, GMVD comprises $53$ multi-view video sequences and $5995$ frames.
Each multi-view video sequence is captured at a frame rate of $2$ fps, and each view video is recorded at $1080 \times 1920$ resolution.
A total of $2800$ pedestrians are contained in GMVD, and each sequence contains $20 \text{--} 40$ pedestrians in each frame on average.
Each scene covers a different size of the region and quantizes the region into a grid using square grid cells of $2.5 \, \mathrm{cm^2}$.
Each grid cell of each sequence is captured by $2.8 \text{--} 6.4$ cameras on average.
Details of the statistics for each scene are provided in Table~\ref{table:data_statistics}.
GMVD splits them into $6$ scenes with $43$ sequences and $4983$ frames for training and $1$ scene (the bottom row in Table~\ref{table:data_statistics}) with $10$ sequences and $1012$ frames for testing.
To make testing difficult, this testing split also contains two different camera layouts: one with $6$ cameras and the other with $8$ cameras.

\section{Implementation Details}
\label{sup:implementation}
During training, we applied random resizing and cropping to multi-view image sequences as the data augmentation following \cite{hou2021multiview, teepe2024earlybird}.
We applied the same augmentation to all multi-view images in one multi-view image sequence within the window.
We set the scale range of the resizing and cropping to $[0.8, 1.2]$.
We also applied dropout at the rate of $0.1$ to the attention layer.
For the Adam optimizer~\cite{loshchilov2018decoupled}, we set the optimizer momentum to $\beta_1 = 0.9$ and $\beta_2 = 0.999$.
We did not use the weight decay.
We set the radius of the Gaussian kernel to $8$ pixels when generating smoothed ground truth occupancy maps.

\begin{table}[t]
\centering
\vspace{-1.0mm}
\scalebox{0.85}{
\begin{tabular}{c|cccc} \hline
Method & MODA$\uparrow$ & MODP$\uparrow$ & Recall$\uparrow$ & Precision$\uparrow$ \\ \hline
EarlyBird~\cite{teepe2024earlybird} & 73.2 & 78.5 & 77.4  & \textbf{94.9} \\
MVFlow~\cite{engilberge2023multi} & 72.4 & 78.2 & 77.7 & 93.5 \\
TrackTacular~\cite{teepe2024lifting} & 68.0 & \textbf{79.3} & 72.9 & 93.7 \\
MVTr~\cite{yang2024end} & 72.7 & 78.9 & 79.0 & 92.6 \\
Ours & \textbf{74.4} & 79.2 & \textbf{79.4} & 94.0 \\ \hline
\end{tabular}
}
\vspace{-2.0mm}
\caption{Comparison of detection results with previous end-to-end MVPT methods on GMVD. 
}
\vspace{-2.0mm}
\label{table:detection}
\end{table}

\begin{table}[t]
\centering
\vspace{-1.0mm}
\scalebox{0.83}{
\begin{tabular}{c|cccc} \hline
 & MODA$\uparrow$ & MODP$\uparrow$ & Recall$\uparrow$ & Precision$\uparrow$ \\ \hline
MVFP~\cite{aung2024enhancing} & 73.3 & 76.5 & 79.2 & 93.0 \\ 
Ours w/ $\mathcal{L}_\mathrm{det}$ only & 71.6 & 78.0 & 78.2 & 92.3 \\ 
Ours w/ $\mathcal{L}_\mathrm{all}$ & \textbf{74.4} & \textbf{79.2} & \textbf{79.4} & \textbf{94.0} \\ \hline
\end{tabular}
}
\vspace{-2.0mm}
\caption{Effect of $\mathcal{L}_\mathrm{all}$ on detection performance and comparison with the state-of-the-art MVPD method. 
}
\vspace{-2.0mm}
\label{table:l_asso}
\end{table}

\section{Analysis of Detection Results}
\label{sup:detection}

In this section, we evaluated and analyzed the detection results of MVTrajecter.
Unless otherwise stated, all experiments were conducted on GMVD.

\begin{figure}[t]
    \centering
    \includegraphics[width=75mm]{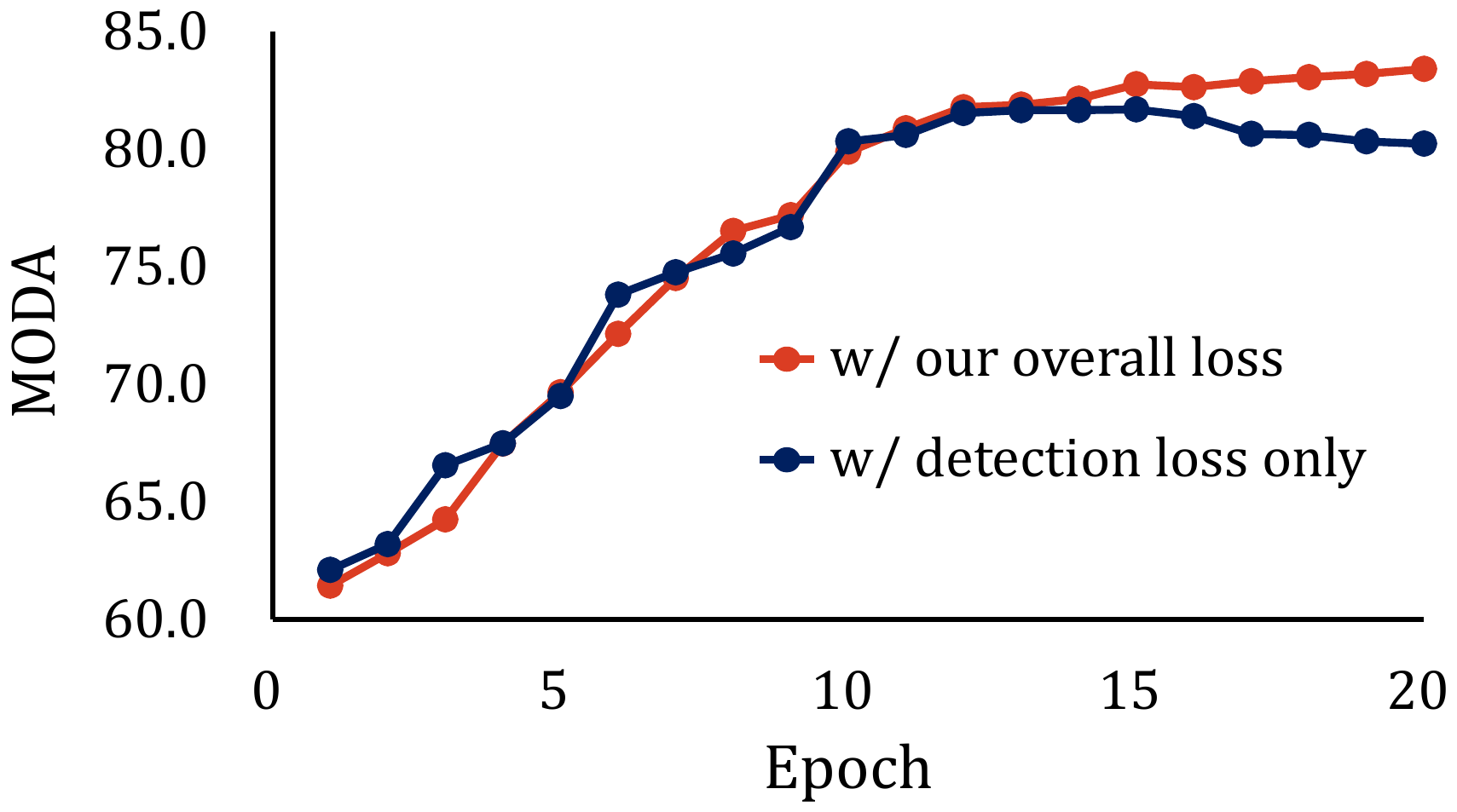}
    \vspace{-2mm}
    \caption{MODA on the validation data of models trained with our overall loss $\mathcal{L}_\mathrm{all}$ and with the detection loss $\mathcal{L}_\mathrm{det}$ only.}
    \vspace{-2mm}
    \label{fig:moda}
\end{figure}

\begin{figure*}[tb]
    \centering
    \includegraphics[width=170mm]{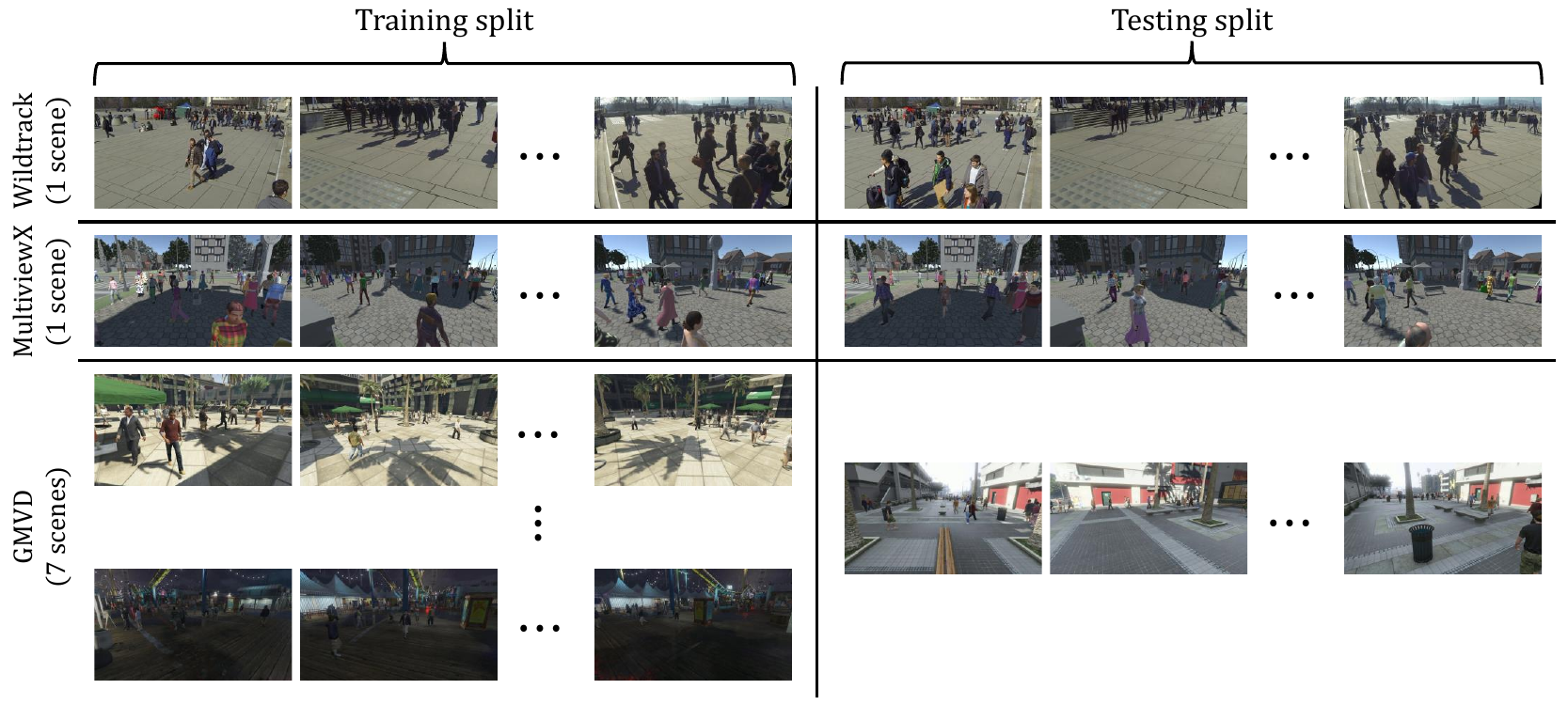}
    \caption{Sample frames of Wildtrack~\cite{chavdarova2018wildtrack}, MultiviewX~\cite{hou2020multiview}, and GMVD~\cite{vora2023bringing}. Top to bottom rows represent Wildtrack, MultiviewX, and GMVD, respectively. Left and right columns show the training split and testing split of each dataset, respectively. While GMVD contains $7$ scenes, we visualize only $3$ of them here.}
    \label{fig:dataset}
\end{figure*}

\paragraph{Evaluation Metrics.}
Following previous studies~\cite{hou2020multiview,vora2023bringing,teepe2024earlybird}, we used four standard metrics provided by Chavdarova~et al.~\cite{chavdarova2018wildtrack} and Kasturi~et al.~\cite{kasturi2008framework}: Multiple Object Detection Accuracy (MODA), Multiple Object Detection Precision (MODP), recall, and precision.
A detected pedestrian was classified as a true positive if its distance from the ground truth was within $0.5$ meters.
We used MODA as the primary performance indicator following previous studies~\cite{hou2020multiview,hou2021multiview,vora2023bringing,teepe2024earlybird,teepe2024lifting}.

\paragraph{Comparison of detection results.}
We compared the detection results of MVTrajecter with EarlyBird~\cite{teepe2024earlybird}, MVFlow~\cite{engilberge2023multi}, TrackTacular~\cite{teepe2024lifting}, and MVTr~\cite{yang2024end}, which are state-of-the-art end-to-end MVPT methods and were compared in Sec.~\ref{sub:result} for their tracking performances.
Table~\ref{table:detection} shows the comparison results.
MVTrajecter achieved the best MODA and recall and the second-best MODP and precision.
This indicates that MVTrajecter is superior to the other methods in terms of detection.
Compared to the other methods, our overall loss $\mathcal{L}_\mathrm{all}$ imposed large temporal constraints on MVTrajecter, which is presumably what suppressed the overfitting in the detection and improved its detection performance.

\paragraph{Effect of $\mathcal{L}_\mathrm{all}$ on the detection.}
We investigated the effect of our overall loss $\mathcal{L}_\mathrm{all}$ on the detection performance.
Table~\ref{table:l_asso} shows a comparison of the detection performance between MVFP~\cite{aung2024enhancing}, which is the state-of-the-art detection method, and MVTrajecter trained with $\mathcal{L}_\mathrm{all}$ and with $\mathcal{L}_\mathrm{det}$ only.
When MVTrajecter was trained only with $\mathcal{L}_\mathrm{det}$, only its detection network was optimized.
Using $\mathcal{L}_\mathrm{all}$ for training greatly improved the detection performance.
In addition, surprisingly, MVTrajecter outperformed MVFP even though MVTrajecter did not utilize the complex projections and detection modules used in MVFP.
Figure~\ref{fig:moda} shows the MODA for the validation data at each epoch of the models trained with $\mathcal{L}_\mathrm{all}$ and with $\mathcal{L}_\mathrm{det}$ only.
While MODA of the model trained only with $\mathcal{L}_\mathrm{det}$ decreased from epoch $12$, MODA of the model trained with $\mathcal{L}_\mathrm{all}$ continuously increased.
This indicates that $\mathcal{L}_\mathrm{all}$ suppressed the overfitting of the detection.

\begin{table}[t]
\centering
\vspace{-1.0mm}
\scalebox{0.85}{
\begin{tabular}{c|ccccc} \hline
 & IDF1$\uparrow$ & MOTA$\uparrow$ & MOTP$\uparrow$ & MT$\uparrow$ & ML$\downarrow$ \\ \hline
Two models & 76.5 & 77.4 & 87.7 & 65.4 & 8.7 \\  
Two branches & \textbf{77.2} & \textbf{78.1} & 87.7 & \textbf{66.0} & 8.7 \\ \hline
\end{tabular}
}
\vspace{-2.0mm}
\caption{Effect of having two branches in one model.
}
\vspace{-1.0mm}
\label{table:branch}
\end{table}

\begin{table}[t]
\centering
\vspace{-1.0mm}
\scalebox{0.93}{
\begin{tabular}{c|ccccc} \hline
$\alpha$ & IDF1$\uparrow$ & MOTA$\uparrow$ & MOTP$\uparrow$ & MT$\uparrow$ & ML$\downarrow$ \\ \hline
$0.97$ & 73.9 & 73.3 & \textbf{90.4} & 61.8 & 0.0 \\  
$0.98$ & \textbf{74.8} & \textbf{75.8} & 90.3 & 61.8 & 0.0 \\ 
$0.99$ & 74.6 & \textbf{75.8} & 90.1 & 61.8 & 0.0 \\ \hline
\end{tabular}
}
\vspace{-2.0mm}
\caption{Impact of weighting parameter $\alpha$ on tracking performances for the validation data.
}
\vspace{-1.0mm}
\label{table:alpha}
\end{table}

\section{Additional Ablation Study}
\label{sup:ablation}

In this section, we conducted ablation studies to investigate the effect of having two branches in one model, the impact of weighting parameter $\alpha$, the value ranges of $C_\mathrm{TMC}$ and $C_\mathrm{TAC}$, the comparison of TAC with EMA aggregation, and the pooling choice.
Unless otherwise stated, all experiments were conducted on GMVD.

\paragraph{Advantage of having two branches.}
MVTrajecter has the motion and appearance branches in one model, as shown in Fig.~\ref{fig:fig2}.
To verify the effectiveness of having two branches in one model, we compared it with the case of tracking performed by two models (\ie, one has the motion branch and the other has the appearance branch).
According to the comparison results in Table~\ref{table:branch}, having two branches achieved a slightly better performance.
This indicates the advantage of simultaneously modeling both motion and appearance in an end-to-end manner.

\begin{table}[t]
\centering
\vspace{-1.0mm}
\scalebox{0.93}{
\begin{tabular}{c|ccccc} \hline
 & IDF1$\uparrow$ & MOTA$\uparrow$ & MOTP$\uparrow$ & MT$\uparrow$ & ML$\downarrow$ \\ \hline
EMA & 74.8 & 76.1 & 87.0 & 64.2 & 8.9 \\  
TAC & \textbf{77.2} & \textbf{78.1} & \textbf{87.7} & \textbf{66.0} & \textbf{8.7} \\ \hline
\end{tabular}
}
\vspace{-2.0mm}
\caption{Comparison between TAC and the exponential moving average (EMA) appearance feature aggregation.
}
\vspace{-1.0mm}
\label{table:ema}
\end{table}

\begin{table}[t]
\centering
\vspace{-1.0mm}
\scalebox{0.93}{
\begin{tabular}{c|ccccc} \hline
Pooling & IDF1$\uparrow$ & MOTA$\uparrow$ & MOTP$\uparrow$ & MT$\uparrow$ & ML$\downarrow$ \\ \hline
Mean & 76.3 & 77.5 & 87.6 & 65.2 & 8.8 \\  
Max & \textbf{77.2} & \textbf{78.1} & \textbf{87.7} & \textbf{66.0} & \textbf{8.7} \\ \hline
\end{tabular}
}
\vspace{-2.0mm}
\caption{Comparison between mean pooling and max pooling in the detection network.
}
\vspace{-1.0mm}
\label{table:pooling}
\end{table}

\paragraph{Impact of weighting parameter $\alpha$ between TMC and TAC.}
As described in Sec.~\ref{sub:implementation}, we tuned $\alpha$ on the validation data.
Here, we show the impact of $\alpha$ on the tracking performances for the validation data in Table~\ref{table:alpha}.
Increasing $\alpha$ improved MOTA while it worsened MOTP.
We therefore set $\alpha = 0.98$ on the basis of these results.

\paragraph{Value ranges of $C_\mathrm{TMC}$ and $C_\mathrm{TAC}$.}
We investigated the value ranges of $C_\mathrm{TMC}$ and $C_\mathrm{TAC}$ during inference to verify the validity of the extreme weighting of $\alpha$.
On GMVD test split and for $K=7$, the value range of $C_\mathrm{TMC}$ was from $0.3$ to $1.2 \times 10^{3}$, and the value range of $C_\mathrm{TAC}$ was from $-6.2$ to $-4.7 \times 10^{-25}$.
Because the value ranges of $C_\mathrm{TMC}$ and $C_\mathrm{TAC}$ differed greatly, the weighted sum using $\alpha=0.98$ worked effectively.

\paragraph{Comparison between TAC and EMA appearance feature aggregation.}
We aggregated appearance features over multiple past timestamps by calculating the probabilities for each timestamp (Eq.~\ref{formula:prob}) and summing the probabilities for all timestamps (Eq.~\ref{formula:tac}) in TAC.
In contrast to our approach, some monocular tracking methods aggregate the appearance features by updating the latest appearance features using the exponential moving average (EMA)~\cite{wang2020towards,yu2022towards,du2021giaotracker,du2023strongsort,aharon2022bot}.
We compared TAC with EMA appearance feature aggregation, and the results shown in Table~\ref{table:ema} indicate that leveraging TAC outperformed leveraging EMA aggregation.
Since EMA aggregation directly fuses the appearance features of pedestrians recognized as identical, we presume that it is more affected by association errors than TAC.

\begin{table}[t]
\centering
\vspace{-1.0mm}
\scalebox{0.93}{\begin{tabular}{c|ccc} \hline
Method & IDF1$\uparrow$ & MOTA$\uparrow$ & MOTP$\uparrow$ \\ \hline
w/ PE in motion branch & 76.9 & 78.0 & \textbf{87.9} \\ 
w/ PE in appearance branch & 76.7 & 77.9 & 87.6 \\ 
w/o PE (Ours) & \textbf{77.2} & \textbf{78.1} & 87.7 \\ \hline
\end{tabular}
}
\vspace{-2.0mm}
\caption{Effect of positional embeddings. ``PE'' means positional embeddings.}
\vspace{-1.0mm}
\label{table:pos}
\end{table}

\begin{table}[t]
\centering
\vspace{-1.0mm}
{
\begin{tabular}{c|c} \hline
Method & HOTA$\uparrow$  \\ \hline
EarlyBird~\cite{teepe2024earlybird} & 64.5 \\ 
MVFlow~\cite{engilberge2023multi} & 66.1 \\ 
TrackTacular~\cite{teepe2024lifting} & 68.2 \\
MVTr~\cite{yang2024end} & 70.7 \\
Ours & \textbf{75.0} \\ \hline
\end{tabular}
}
\vspace{-2.0mm}
\caption{Comparison with previous methods on the HOTA metric when all models used their own detection results for tracking.}
\vspace{-1.0mm}
\label{table:hota}
\end{table}

\paragraph{Pooling choice.}
We performed max pooling to aggregate projected features from multiple views, as described in Sec.~\ref{sub:architecture}.
Another option is to perform mean pooling instead of max pooling.
When we compared these two pooling operations, as shown in Table~\ref{table:pooling}, max pooling achieved a better tracking performance.
Max pooling extracts the most relevant and informative features from different view perspectives for the subsequent modules, which is presumably what resulted in better tracking performance.

\paragraph{Effect of positional embedding.}
While we used temporal embeddings (see Sec.~\ref{sub:architecture}), we did not use positional embeddings because the BEV features implicitly contain the positional information.
To justify this, we implemented learnable positional embeddings in the motion and appearance branches, as shown in Table~\ref{table:pos}.
We did not observe the improvement from our original implementation.
Therefore, positional embeddings are not crucial for the motion and appearance branches.

\paragraph{Evaluation on HOTA metric.}

While we did not use Higher Order Tracking Accuracy (HOTA)~\cite{luiten2021hota} because comparison methods were not evaluated on it in their original papers, it is one of the important evaluation metrics in the field of monocular tracking.
Therefore, we compared the models in Table~\ref{table:gmvd_det_track} using the HOTA metric.
Table~\ref{table:hota} shows the comparison results.
Our method also significantly outperformed previous methods on the HOTA metric.
This also demonstrates the effectiveness of our proposed method.

\begin{table}[t]
\centering
\vspace{-1.0mm}
{
\begin{tabular}{c|c} \hline
$K$ & FPS \\ \hline
Detection network & 6.4 \\
$1$ & 5.9 \\ 
$3$ & 5.8 \\ 
$5$ & 5.6 \\
$7$ & 5.4 \\ \hline
\end{tabular}
}
\vspace{-2.0mm}
\caption{Effect of past trajectory length $K$ on inference speed.}
\vspace{-1.0mm}
\label{table:fps_k}
\end{table}

\begin{table}[t]
\centering
\vspace{-1.0mm}
{
\begin{tabular}{c|c} \hline
Method & FPS \\ \hline
EarlyBird~\cite{teepe2024earlybird} & 6.0 \\ 
MVFlow~\cite{engilberge2023multi} & 6.1 \\ 
TrackTacular~\cite{teepe2024lifting} & 6.2 \\
MVTr~\cite{yang2024end} & 6.1 \\
Ours & 5.4 \\ \hline
\end{tabular}
}
\vspace{-2.0mm}
\caption{Comparison of inference speed with previous methods.}
\vspace{-1.0mm}
\label{table:fps_comparison}
\end{table}

\section{Analysis of Inference Speed}
\label{sup:speed}

\paragraph{Effect of past trajectory length $K$.}
Introducing more past information inevitably causes a decrease in inference speed.
We measured the inference speed of the models in Table~\ref{table:length} and the detection network of our method on an Nvidia A100 GPU.
The FPS of the models at $K=1,3,5,7$ and the detection network are shown in Table~\ref{table:fps_k}.
Since the detection network occupies most of the runtime, the effect of increasing $K$ on inference speed is very small.

\paragraph{Comparison with previous methods.}

We also measured the inference speed of the models in Table~\ref{table:gmvd_det_track}.
The FPS of EarlyBird~\cite{teepe2024earlybird}, MVFlow~\cite{engilberge2023multi}, TrackTacular~\cite{teepe2024lifting}, MVTr~\cite{yang2024end}, and ours are shown in Table~\ref{table:fps_comparison}.
Our method is slightly slower than the others because it uses more past information.
However, this speed gap is very small compared to the tracking performance improvement.

\section{Qualitative Results}
\label{sup:qualiative}\

\begin{figure*}[tb]
    \centering
    \includegraphics[width=160mm]{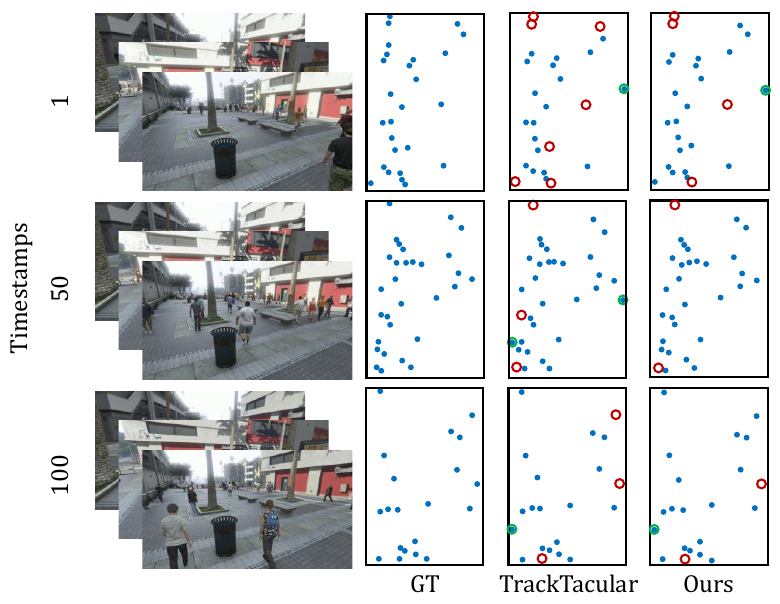}
    \caption{Qualitative comparison of detection results between ground truth (GT), TrackTacular~\cite{teepe2024lifting}, and the proposed MVTrajecter on GMVD. Blue filled circles represent detected pedestrians, red circles represent missed detections, and green circles represent false positives.}
    \label{fig:qualitative_gmvd_detection}
\end{figure*}

\begin{figure*}[tb]
    \centering
    \includegraphics[width=170mm]{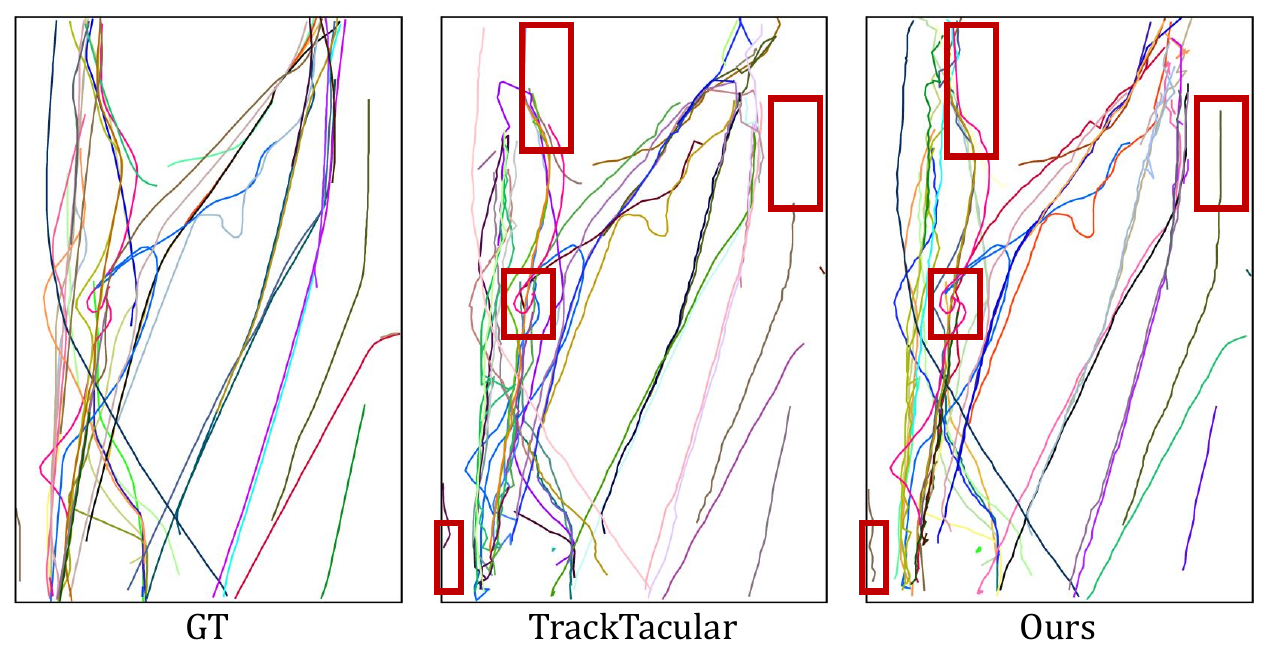}
    \caption{Qualitative comparison of tracking results between ground truth (GT), TrackTacular~\cite{teepe2024lifting}, and the proposed MVTrajecter on GMVD. Each line represents a pedestrian track. Red squares indicate examples of different results between TrackTacular and our MVTrajecter.}
    \label{fig:qualitative_gmvd_track}
\end{figure*}

To visually verify the effectiveness of our MVTrajecter, we visualized the detection and tracking results of MVTrajecter, TrackTacular~\cite{teepe2024lifting}, and the ground truth.
Figures~\ref{fig:qualitative_gmvd_detection} and \ref{fig:qualitative_gmvd_track} show the comparison of the detection and tracking results on one multi-view video sequence (consisting of $100$ timestamps) of GMVD, respectively.
In Fig.~\ref{fig:qualitative_gmvd_detection}, our MVTrajecter reduced the number of missed detections and false positive detections, which are denoted by red and green circles, respectively.
This demonstrates that MVTrajecter is superior to TrackTacular in the detection.
In Fig.~\ref{fig:qualitative_gmvd_track}, we can see that MVTrajecter correctly tracked pedestrians that TrackTacular failed to track, as indicated by the red squares.
This demonstrates that MVTrajecter is also superior to TrackTacular in tracking.
While MVTrajecter improved the detection and tracking performance, it still struggled to detect and track pedestrians near the boundaries of the occupancy maps.
Therefore, in order to improve performance further, methods that can accurately handle pedestrians near the boundaries are needed.

\begin{figure*}[tb]
    \centering
    \includegraphics[width=120mm]{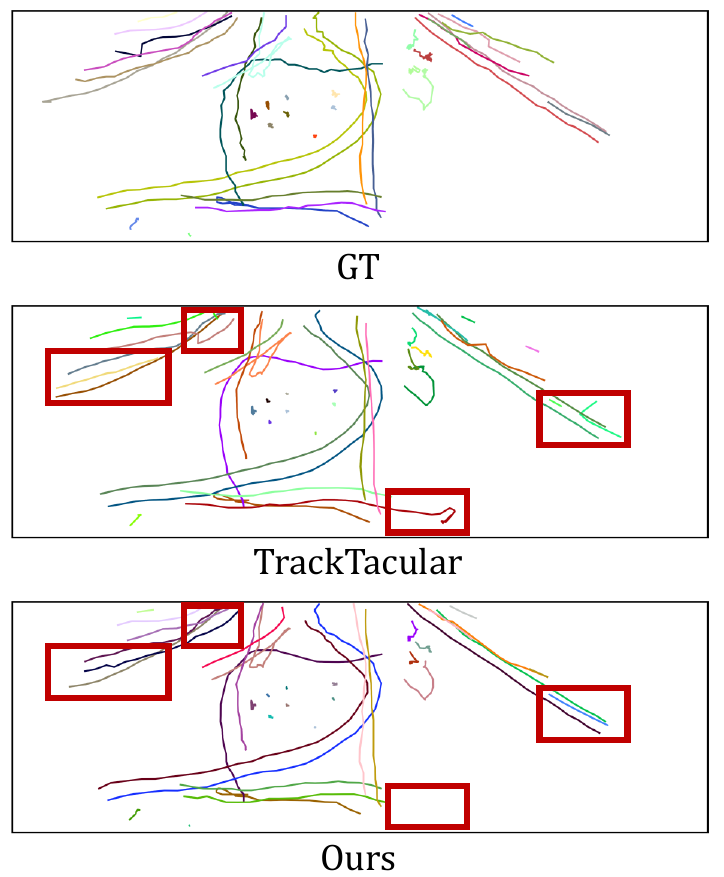}
    \caption{Qualitative comparison of tracking results between ground truth (GT), TrackTacular~\cite{teepe2024lifting}, and the proposed MVTrajecter on Wildtrack. Each line represents a pedestrian track. Red squares indicate examples of different results between TrackTacular and our MVTrajecter.}
    \label{fig:qualitative_wildtrack}
\end{figure*}

\begin{figure*}[tb]
    \centering
    \includegraphics[width=90mm]{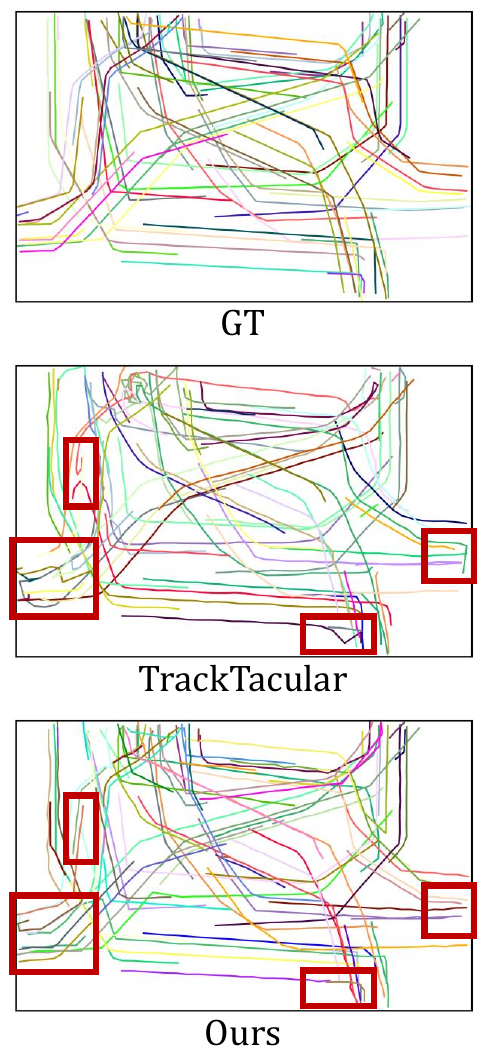}
    \caption{Qualitative comparison of tracking results between ground truth (GT), TrackTacular~\cite{teepe2024lifting}, and the proposed MVTrajecter on MultiviewX. Each line represents a pedestrian track. Red squares indicate examples of different results between TrackTacular and our MVTrajecter.}
    \label{fig:qualitative_multiviewx}
\end{figure*}

To validate the versatility of MVTrajecter, which is not limited to GMVD, we also visualized tracking results on Wildtrack and MultiviewX.
Figures~\ref{fig:qualitative_wildtrack} and \ref{fig:qualitative_multiviewx} show the comparison of tracking results on Wildtrack and MultiviewX, respectively.
In both figures, we can see that even in cases where TrackTacular failed to track pedestrians, MVTrajecter correctly tracked them, as indicated by the red squares.
These results demonstrate that MVTrajecter is effective for various datasets.

\end{document}